\documentclass[10pt,twocolumn,letterpaper]{article}

\usepackage{cvpr}
\usepackage{times}
\usepackage{epsfig}
\usepackage{graphicx}
\usepackage{amsmath}
\usepackage{amssymb}
\usepackage{color}
\usepackage{booktabs,threeparttable}
\usepackage{colortbl}
\usepackage[export]{adjustbox}
\usepackage{caption}

\usepackage[pagebackref=true,breaklinks=true,letterpaper=true,colorlinks,bookmarks=false]{hyperref}

\cvprfinalcopy 

\def\cvprPaperID{} 

\begin{document}

\title{YOLOv4: Optimal Speed and Accuracy of Object Detection}

\author{Alexey Bochkovskiy$^*$\\
{\tt\small alexeyab84@gmail.com}
\and
Chien-Yao Wang$^*$\\
Institute of Information Science\\
Academia Sinica, Taiwan\\
{\tt\small kinyiu@iis.sinica.edu.tw}
\and
Hong-Yuan Mark Liao\\
Institute of Information Science\\
Academia Sinica, Taiwan\\
{\tt\small liao@iis.sinica.edu.tw}
}

\maketitle

\begin{abstract}
  There are a huge number of features which are said to improve Convolutional Neural Network (CNN) accuracy. Practical testing of combinations of such features on large datasets, and theoretical justification of the result, is required. Some features operate on certain models exclusively and for certain problems exclusively, or only for small-scale datasets; while some features, such as batch-normalization and residual-connections, are applicable to the majority of models, tasks, and datasets. We assume that such universal features include Weighted-Residual-Connections (WRC), Cross-Stage-Partial-connections (CSP), Cross mini-Batch Normalization (CmBN), Self-adversarial-training (SAT) and Mish-activation.  We use new features: WRC, CSP, CmBN, SAT, Mish activation, Mosaic data augmentation, CmBN, DropBlock regularization, and CIoU loss, and combine some of them to achieve state-of-the-art results: 43.5\% AP (65.7\% AP$_{50}$) for the MS COCO dataset at a real-time speed of $\sim$65 FPS on Tesla V100. Source code is at \url{https://github.com/AlexeyAB/darknet}.

\end{abstract}


\section{Introduction}

The majority of CNN-based object detectors are largely applicable only for recommendation systems. For example, searching for free parking spaces via urban video cameras is executed by slow accurate models, whereas car collision warning is related to fast inaccurate models. Improving the real-time object detector accuracy enables using them not only for hint generating recommendation systems, but also for stand-alone process management and human input reduction. Real-time object detector operation on conventional Graphics Processing Units (GPU) allows their mass usage at an affordable price. The most accurate modern neural networks do not operate in real time and require large number of GPUs for training with a large mini-batch-size. We address such problems through creating a CNN that operates in real-time on a conventional GPU, and for which training requires only one conventional GPU.

\begin{figure}[t]
	\begin{center}
		\includegraphics[width=1.0\linewidth]{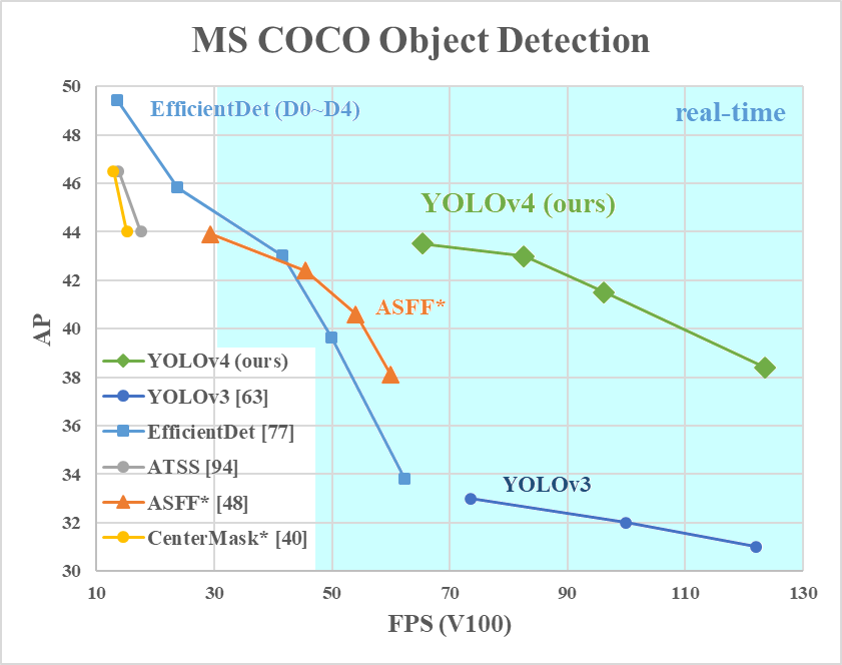}
	\end{center}
    \vspace{-4mm}
	\caption{Comparison of the proposed YOLOv4 and other state-of-the-art object detectors. YOLOv4 runs twice faster than EfficientDet with comparable performance. Improves YOLOv3's AP and FPS by 10\% and 12\%, respectively.}
	\label{fig:ap}
\end{figure}

The main goal of this work is designing a fast operating speed of an object detector in production systems and optimization for parallel computations, rather than the low computation volume theoretical indicator (BFLOP). We hope that the designed object can be easily trained and used. For example, anyone who uses a conventional GPU to train and test can achieve real-time, high quality, and convincing object detection results, as the YOLOv4 results shown in  Figure \ref{fig:ap}. Our contributions are summarized as follows:

\begin{enumerate}
	\item We develope an efficient and powerful object detection model. It makes everyone can use a 1080 Ti or 2080 Ti GPU to train a super fast and accurate object detector.
	\item We verify the influence of state-of-the-art Bag-of-Freebies and Bag-of-Specials methods of object detection during the detector training.
	\item We modify state-of-the-art methods and make them more effecient and suitable for single GPU training, including CBN \cite{yao2020cross}, PAN \cite{liu2018path}, SAM \cite{woo2018cbam}, etc.
\end{enumerate}

\vspace{-2mm}



\begin{figure*}[t]
	\begin{center}
		\includegraphics[width=0.99\linewidth]{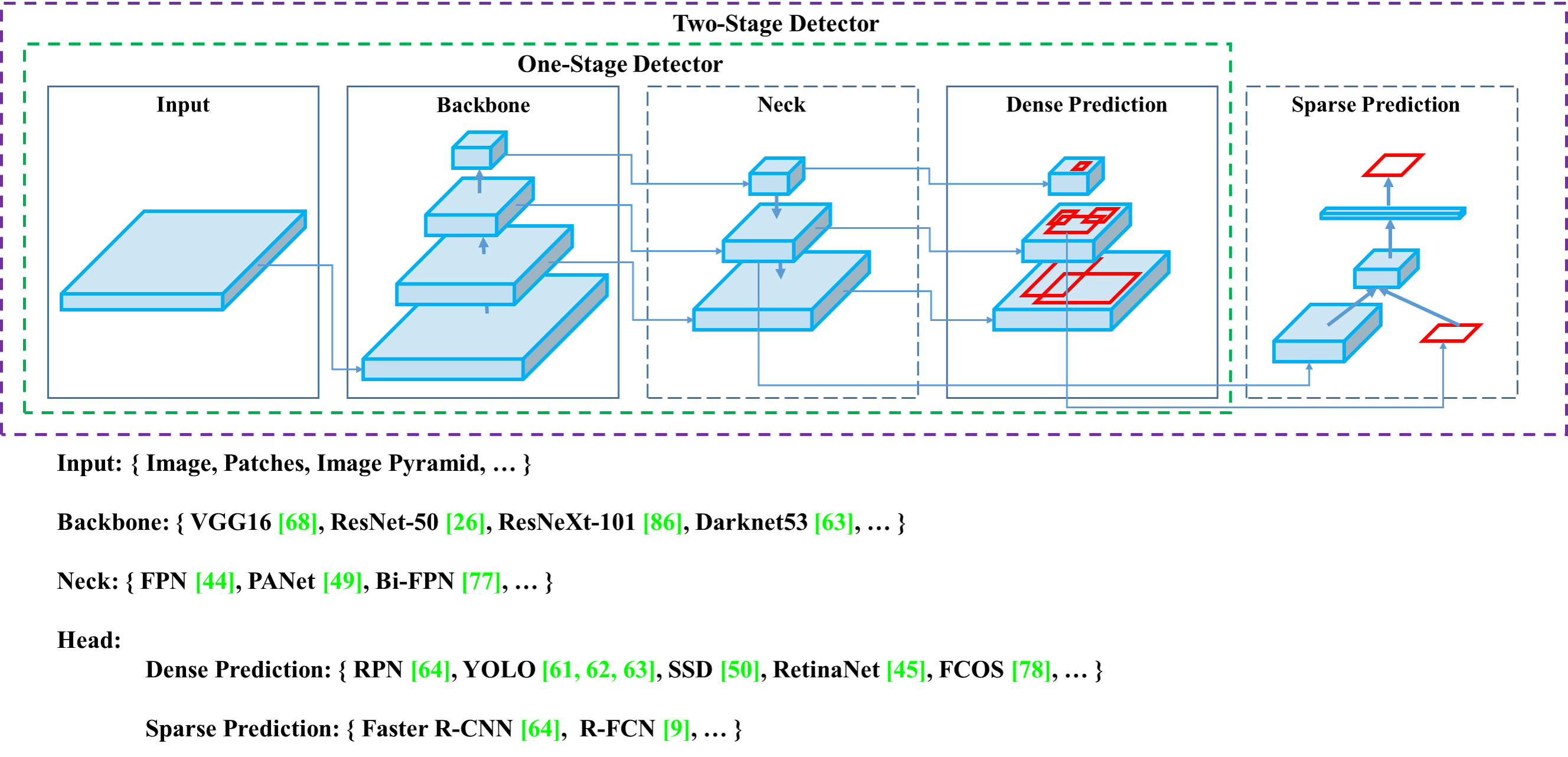}
	\end{center}
    \vspace{-4mm}
	\caption{Object detector.  }
	\label{fig:detector}
\end{figure*}

\section{Related work}

\subsection{Object detection models}

A modern detector is usually composed of two parts, a backbone which is pre-trained on ImageNet and a head which is used to predict classes and bounding boxes of objects. For those detectors running on GPU platform, their backbone could be VGG \cite{simonyan2014very}, ResNet \cite{he2016deep}, ResNeXt \cite{xie2017aggregated}, or DenseNet \cite{huang2017densely}. For those detectors running on CPU platform, their backbone could be SqueezeNet \cite{iandola2016squeezenet}, MobileNet \cite{howard2017mobilenets, sandler2018mobilenetv2, howard2019searching, tan2019mnasnet}, or ShuffleNet \cite{zhang2018shufflenet, ma2018shufflenetv2}. As to the head part, it is usually categorized into two kinds, i.e., one-stage object detector and two-stage object detector. The most representative two-stage object detector is the R-CNN \cite{girshick2014rich} series, including fast R-CNN \cite{girshick2015fast}, faster R-CNN \cite{ren2015faster}, R-FCN \cite{dai2016r}, and Libra R-CNN \cite{pang2019libra}. It is also possible to make a two-stage object detector an anchor-free object detector, such as RepPoints \cite{yang2019reppoints}.  As for one-stage object detector, the most representative models are YOLO \cite{redmon2016you, redmon2017yolo9000, redmon2018yolov3}, SSD \cite{liu2016ssd}, and RetinaNet \cite{lin2017focal}. In recent years, anchor-free one-stage object detectors are developed. The detectors of this sort are CenterNet \cite{duan2019centernet}, CornerNet \cite{law2018cornernet, law2019cornernet}, FCOS \cite{tian2019fcos}, etc. Object detectors developed in recent years often insert some layers between backbone and head, and these layers are usually used to collect feature maps from different stages. We can call it “the neck of an object detector.” Usually, a neck is composed of several bottom-up paths and several top-down paths. Networks equipped with this mechanism include Feature Pyramid Network (FPN) \cite{lin2017feature}, Path Aggregation Network (PAN) \cite{liu2018path}, BiFPN \cite{tan2019efficientdet}, and NAS-FPN \cite{ghiasi2019fpn}. In addition to the above models, some researchers put their emphasis on directly building a new backbone (DetNet \cite{li2018detnet}, DetNAS \cite{chen2019detnas}) or a new whole model (SpineNet \cite{du2019spinenet}, HitDetector \cite{guo2020hit}) for object detection.

To sum up, an ordinary object detector is composed of several parts:

\begin{itemize}
	\item \textbf{Input}: Image, Patches, Image Pyramid
	\item \textbf{Backbones}: VGG16 \cite{simonyan2014very}, ResNet-50 \cite{he2016deep}, SpineNet \cite{du2019spinenet}, EfficientNet-B0/B7 \cite{tan2019efficientnet}, CSPResNeXt50 \cite{wang2020cspnet}, CSPDarknet53 \cite{wang2020cspnet}
	\item \textbf{Neck}:
	\begin{itemize}
		\item[$\bullet$] \textbf{Additional blocks}:  SPP \cite{he2015spatial}, ASPP \cite{chen2017deeplab}, RFB \cite{liu2018receptive}, SAM \cite{woo2018cbam}
		\item[$\bullet$] \textbf{Path-aggregation blocks}: FPN \cite{lin2017feature}, PAN \cite{liu2018path}, NAS-FPN \cite{ghiasi2019fpn}, Fully-connected FPN, BiFPN \cite{tan2019efficientdet}, ASFF \cite{liu2019learning}, SFAM \cite{zhao2019m2det}
	\end{itemize}
	\item \textbf{Heads:}:
	\begin{itemize}
		\item[$\bullet$] \textbf{Dense Prediction (one-stage)}:    	
		\begin{itemize}
			\item[$\circ$] RPN \cite{ren2015faster}, SSD \cite{liu2016ssd}, YOLO \cite{redmon2016you}, RetinaNet \cite{lin2017focal} (anchor based)
			\item[$\circ$] CornerNet \cite{law2018cornernet}, CenterNet \cite{duan2019centernet}, MatrixNet \cite{rashwan2019matrix}, FCOS \cite{tian2019fcos} (anchor free)
		\end{itemize}
		\item[$\bullet$] \textbf{Sparse Prediction (two-stage)}: 
		\begin{itemize}
			\item[$\circ$] Faster R-CNN \cite{ren2015faster}, R-FCN \cite{dai2016r}, Mask R-CNN \cite{he2017mask} (anchor based)
			\item[$\circ$] RepPoints \cite{yang2019reppoints} (anchor free)
		\end{itemize}
	\end{itemize}
\end{itemize}

\subsection{Bag of freebies}

Usually, a conventional object detector is trained off-line. Therefore, researchers always like to take this advantage and develop better training methods which can make the object detector receive better accuracy without increasing the inference cost. We call these methods that only change the training strategy or only increase the training cost as ``bag of freebies.'' What is often adopted by object detection methods and meets the definition of bag of freebies is data augmentation. The purpose of data augmentation is to increase the variability of the input images, so that the designed object detection model has higher robustness to the images obtained from different environments. For examples, photometric distortions and geometric distortions are two commonly used data augmentation method and they definitely benefit the object detection task. In dealing with photometric distortion, we adjust the brightness, contrast, hue, saturation, and noise of an image. For geometric distortion, we add random scaling, cropping, flipping, and rotating.

The data augmentation methods mentioned above are all pixel-wise adjustments, and all original pixel information in the adjusted area is retained. In addition, some researchers engaged in data augmentation put their emphasis on simulating object occlusion issues. They have achieved good results in image classification and object detection. For example, random erase \cite{zhong2017random} and CutOut \cite{devries2017improved} can randomly select the rectangle region in an image and fill in a random or complementary value of zero. As for hide-and-seek \cite{singh2018hide} and grid mask \cite{chen2020gridmask}, they randomly or evenly select multiple rectangle regions in an image and replace them to all zeros. If similar concepts are applied to feature maps, there are DropOut \cite{srivastava2014dropout}, DropConnect \cite{wan2013regularization}, and DropBlock \cite{ghiasi2018dropblock} methods. In addition, some researchers have proposed the methods of using multiple images together to perform data augmentation. For example, MixUp \cite{zhang2017mixup} uses two images to multiply and superimpose with different coefficient ratios, and then adjusts the label with these superimposed ratios. As for CutMix \cite{yun2019cutmix}, it is to cover the cropped image to rectangle region of other images, and adjusts the label according to the size of the mix area. In addition to the above mentioned methods, style transfer GAN \cite{geirhos2018imagenet} is also used for data augmentation, and such usage can effectively reduce the texture bias learned by CNN.

Different from the various approaches proposed above, some other bag of freebies methods are dedicated to solving the problem that the semantic distribution in the dataset may have bias. In dealing with the problem of semantic distribution bias, a very important issue is that there is a problem of data imbalance between different classes, and this problem is often solved by hard negative example mining \cite{sung1998example} or online hard example mining \cite{shrivastava2016training} in two-stage object detector. But the example mining method is not applicable to one-stage object detector, because this kind of detector belongs to the dense prediction architecture. Therefore Lin \etal. \cite{lin2017focal} proposed focal loss to deal with the problem of data imbalance existing between various classes. Another very important issue is that it is difficult to express the relationship of the degree of association between different categories with the one-hot hard representation. This representation scheme is often used when executing labeling. The label smoothing proposed in \cite{szegedy2016rethinking} is to convert hard label into soft label for training, which can make model more robust. In order to obtain a better soft label, Islam \etal. \cite{islam2017label} introduced the concept of knowledge distillation to design the label refinement network.

The last bag of freebies is the objective function of Bounding Box (BBox) regression. The traditional object detector usually uses Mean Square Error (MSE) to directly perform regression on the center point coordinates and height and width of the BBox, i.e., \{$x_{center}$, $y_{center}$, $w$, $h$\}, or the upper left point and the lower right point, i.e., \{$x_{top\_left}$, $y_{top\_left}$, $x_{bottom\_right}$, $y_{bottom\_right}$\}. As for anchor-based method, it is to estimate the corresponding offset, for example \{$x_{center\_offset}$, $y_{center\_offset}$, $w_{offset}$, $h_{offset}$\} and  \{$x_{top\_left\_offset}$, $y_{top\_left\_offset}$, $x_{bottom\_right\_offset}$, $y_{bottom\_right\_offset}$\}. However, to directly estimate the coordinate values of each point of the BBox is to treat these points as independent variables, but in fact does not consider the integrity of the object itself. In order to make this issue processed better, some researchers recently proposed IoU loss \cite{yu2016unitbox}, which puts the coverage of predicted BBox area and ground truth BBox area into consideration. The IoU loss computing process will trigger the calculation of the four coordinate points of the BBox by executing IoU with the ground truth, and then connecting the generated results into a whole code. Because IoU is a scale invariant representation, it can solve the problem that when traditional methods calculate the $l_1$ or $l_2$ loss of \{$x$, $y$, $w$, $h$\}, the loss will increase with the scale. Recently, some researchers have continued to improve IoU loss. For example, GIoU loss \cite{rezatofighi2019generalized} is to include the shape and orientation of object in addition to the coverage area. They proposed to find the smallest area BBox that can simultaneously cover the predicted BBox and ground truth BBox, and use this BBox as the denominator to replace the denominator originally used in IoU loss. As for DIoU loss \cite{zheng2019distance}, it additionally considers the distance of the center of an object, and CIoU loss \cite{zheng2019distance}, on the other hand simultaneously considers the overlapping area, the distance between center points, and the aspect ratio. CIoU can achieve better convergence speed and accuracy on the BBox regression problem.

\newpage

\subsection{Bag of specials}

For those plugin modules and post-processing methods that only increase the inference cost by a small amount but can significantly improve the accuracy of object detection, we call them ``bag of specials''. Generally speaking, these plugin modules are for enhancing certain attributes in a model, such as enlarging receptive field, introducing attention mechanism, or strengthening feature integration capability, etc., and post-processing is a method for screening model prediction results.

Common modules that can be used to enhance receptive field are SPP \cite{he2015spatial}, ASPP \cite{chen2017deeplab}, and RFB \cite{liu2018receptive}. The SPP module was originated from Spatial Pyramid Matching (SPM) \cite{lazebnik2006beyond}, and SPM’s original method was to split feature map into several $d \times d$ equal blocks, where $d$ can be $\{1, 2, 3, ...\}$, thus forming spatial pyramid, and then extracting bag-of-word features. SPP integrates SPM into CNN and use max-pooling operation instead of bag-of-word operation. Since the SPP module proposed by He \etal. \cite{he2015spatial} will output one dimensional feature vector, it is infeasible to be applied in Fully Convolutional Network (FCN). Thus in the design of YOLOv3 \cite{redmon2018yolov3}, Redmon and Farhadi improve SPP module to the concatenation of max-pooling outputs with kernel size $k \times k$, where $k = \{1, 5, 9, 13\}$, and stride equals to 1. Under this design, a relatively large $k \times k$ max-pooling effectively increase the receptive field of backbone feature. After adding the improved version of SPP module, YOLOv3-608 upgrades AP$_{50}$ by 2.7\% on the MS COCO object detection task at the cost of 0.5\% extra computation. The difference in operation between ASPP \cite{chen2017deeplab} module and improved SPP module is mainly from the original $k \times k$ kernel size, max-pooling of stride equals to 1 to several $3 \times 3$ kernel size, dilated ratio equals to $k$, and stride equals to 1 in dilated convolution operation. RFB module is to use several dilated convolutions of $k \times k$ kernel, dilated ratio equals to $k$, and stride equals to 1 to obtain a more comprehensive spatial coverage than ASPP. RFB \cite{liu2018receptive} only costs 7\% extra inference time to increase the AP$_{50}$ of SSD on MS COCO by 5.7\%.

The attention module that is often used in object detection is mainly divided into channel-wise attention and point-wise attention, and the representatives of these two attention models are Squeeze-and-Excitation (SE) \cite{hu2018squeeze} and Spatial Attention Module (SAM) \cite{woo2018cbam}, respectively. Although SE module can improve the power of ResNet50 in the ImageNet image classification task 1\% top-1 accuracy at the cost of only increasing the computational effort by 2\%, but on a GPU usually it will increase the inference time by about 10\%, so it is more appropriate to be used in mobile devices. But for SAM, it only needs to pay 0.1\% extra calculation and it can improve ResNet50-SE 0.5\% top-1 accuracy on the ImageNet image classification task. Best of all, it does not affect the speed of inference on the GPU at all.

In terms of feature integration, the early practice is to use skip connection \cite{long2015fully} or hyper-column \cite{hariharan2015hypercolumns} to integrate low-level physical feature to high-level semantic feature. Since multi-scale prediction methods such as FPN have become popular, many lightweight modules that integrate different feature pyramid have been proposed. The modules of this sort include SFAM \cite{zhao2019m2det}, ASFF \cite{liu2019learning}, and BiFPN \cite{tan2019efficientdet}. The main idea of SFAM is to use SE module to execute channel-wise level re-weighting on multi-scale concatenated feature maps. As for ASFF, it uses softmax as point-wise level re-weighting and then adds feature maps of different scales. In BiFPN, the multi-input weighted residual connections is proposed to execute scale-wise level re-weighting, and then add feature maps of different scales.

In the research of deep learning, some people put their focus on searching for good activation function. A good activation function can make the gradient more efficiently propagated, and at the same time it will not cause too much extra computational cost. In 2010, Nair and Hinton \cite{nair2010rectified} propose ReLU to substantially solve the gradient vanish problem which is frequently encountered in traditional tanh and sigmoid activation function. Subsequently, LReLU \cite{maas2013rectifier}, PReLU \cite{he2015delving}, ReLU6 \cite{howard2017mobilenets}, Scaled Exponential Linear Unit (SELU) \cite{klambauer2017self}, Swish \cite{ramachandran2017searching}, hard-Swish \cite{howard2019searching}, and Mish \cite{misra2019mish}, etc., which are also used to solve the gradient vanish problem, have been proposed. The main purpose of LReLU and PReLU is to solve the problem that the gradient of ReLU is zero when the output is less than zero. As for ReLU6 and hard-Swish, they are specially designed for quantization networks. For self-normalizing a neural network, the SELU activation function is proposed to satisfy the goal. One thing to be noted is that both Swish and Mish are continuously differentiable activation function.

The post-processing method commonly used in deep-learning-based object detection is NMS, which can be used to filter those BBoxes that badly predict the same object, and only retain the candidate BBoxes with higher response. The way NMS tries to improve is consistent with the method of optimizing an objective function. The original method proposed by NMS does not consider the context information, so Girshick \etal. \cite{girshick2014rich} added classification confidence score in R-CNN as a reference, and according to the order of confidence score, greedy NMS was performed in the order of high score to low score. As for soft NMS \cite{bodla2017soft}, it considers the problem that the occlusion of an object may cause the degradation of confidence score in greedy NMS with IoU score. The DIoU NMS \cite{zheng2019distance} developer’s way of thinking is to add the information of the center point distance to the BBox screening process on the basis of soft NMS. It is worth mentioning that, since none of above post-processing methods directly refer to the captured image features, post-processing is no longer required in the subsequent development of an anchor-free method.


\begin{table*}[h]
	\centering
	\begin{threeparttable}[h]
		\footnotesize
		\caption{Parameters of neural networks for image classification.}
		\label{table:parameters}
		\begin{tabular}{ccccccc}
			\toprule
			Backbone model & \begin{tabular}{@{}c@{}}Input network \\ resolution\end{tabular} & \begin{tabular}{@{}c@{}}Receptive \\ field size\end{tabular} & Parameters & \begin{tabular}{@{}c@{}@{}}Average size \\ of layer output \\ (WxHxC)\end{tabular} & \begin{tabular}{@{}c@{}}BFLOPs \\ (512x512 network resolution)\end{tabular} & \begin{tabular}{@{}c@{}}FPS \\ (GPU RTX 2070)\end{tabular} \\			
			\midrule	
			CSPResNext50 & 512x512 & 425x425 & 20.6 M & \textbf{1058 K} & 31 
			(15.5 FMA) & 62 \\
			CSPDarknet53 & 512x512 & 725x725 & \textbf{27.6 M} & 950 K & 52 
			(26.0 FMA) & \textbf{66} \\
			EfficientNet-B3 (ours) & 512x512 & \textbf{1311x1311} & 12.0 M & 668 K & 11 
			(5.5 FMA) & 26 \\		
			\bottomrule
		\end{tabular}
	\end{threeparttable}
\end{table*}

\section{Methodology}

The basic aim is fast operating speed of neural network, in production systems and optimization for parallel computations, rather than the low computation volume theoretical indicator (BFLOP). We present two options of real-time neural networks:

\begin{itemize}
	\item For GPU – we use a small number of groups (1 - 8) in convolutional layers: CSPResNeXt50 / CSPDarknet53
	\item For VPU - we use grouped-convolution, but we refrain from using Squeeze-and-excitement (SE) blocks - specifically this includes the following models: EfficientNet-lite / MixNet \cite{tan2019mixnet} / GhostNet \cite{han2019ghostnet} / MobileNetV3
\end{itemize}


\subsection{Selection of architecture}

Our objective is to find the optimal balance among the input network resolution, the convolutional layer number, the parameter number (filter\_size$^2$ * filters * channel / groups), and the number of layer outputs (filters). For instance, our numerous studies demonstrate that the CSPResNext50 is considerably better compared to CSPDarknet53 in terms of object classification on the ILSVRC2012 (ImageNet) dataset \cite{deng2009imagenet}. However, conversely, the CSPDarknet53 is better compared to CSPResNext50 in terms of detecting objects on the MS COCO dataset \cite{lin2014microsoft}.

The next objective is to select additional blocks for increasing the receptive field and the best method of parameter aggregation from different backbone levels for different detector levels: e.g. FPN, PAN, ASFF, BiFPN.

A reference model which is optimal for classification is not always optimal for a detector. In contrast to the classifier, the detector requires the following:

\begin{itemize}
	\item Higher input network size (resolution) -- for detecting multiple small-sized objects
	\item More layers -- for a higher receptive field to cover the increased size of input network
	\item More parameters -- for greater capacity of a model to detect multiple objects of different sizes in a single image
\end{itemize}

Hypothetically speaking, we can assume that a model with a larger receptive field size (with a larger number of convolutional layers $3 \times 3$) and a larger number of parameters should be selected as the backbone. Table \ref{table:parameters} shows the information of CSPResNeXt50, CSPDarknet53, and EfficientNet B3. The CSPResNext50 contains only 16 convolutional layers $3 \times 3$, a $425 \times 425$ receptive field and 20.6 M parameters, while CSPDarknet53 contains 29 convolutional layers $3 \times 3$, a $725 \times 725$ receptive field and 27.6 M parameters. This theoretical justification, together with our numerous experiments, show that CSPDarknet53 neural network is the optimal model of the two as the backbone for a detector.

The influence of the receptive field with different sizes is summarized as follows:

\begin{itemize}
	\item Up to the object size - allows viewing the entire object
	\item Up to network size - allows viewing the context around the object
	\item Exceeding the network size - increases the number of connections between the image point and the final activation
\end{itemize}

We add the SPP block over the CSPDarknet53, since it significantly increases the receptive field, separates out the most significant context features and causes almost no reduction of the network operation speed. We use PANet as the method of parameter aggregation from different backbone levels for different detector levels, instead of the FPN used in YOLOv3.

Finally, we choose CSPDarknet53 backbone, SPP additional module, PANet path-aggregation neck, and YOLOv3 (anchor based) head as the architecture of YOLOv4.

In the future we plan to expand significantly the content of Bag of Freebies (BoF) for the detector, which theoretically can address some problems and increase the detector accuracy, and sequentially check the influence of each feature in an experimental fashion.

We do not use Cross-GPU Batch Normalization (CGBN or SyncBN) or expensive specialized devices. This allows anyone to reproduce our state-of-the-art outcomes on a conventional graphic processor e.g. GTX 1080Ti or RTX 2080Ti.

\subsection{Selection of BoF and BoS}

For improving the object detection training, a CNN usually uses the following:

\begin{itemize}
	\item \textbf{Activations}: ReLU, leaky-ReLU, parametric-ReLU, ReLU6, SELU, Swish, or Mish
	\item \textbf{Bounding box regression loss}: MSE, IoU, GIoU, CIoU, DIoU
	\item \textbf{Data augmentation}: CutOut, MixUp, CutMix
	\item \textbf{Regularization method}: DropOut, DropPath \cite{larsson2016fractalnet}, Spatial DropOut \cite{tompson2015efficient}, or DropBlock
	\item \textbf{Normalization of the network activations by their mean and variance}: Batch Normalization (BN) \cite{ioffe2015batch}, Cross-GPU Batch Normalization (CGBN or SyncBN) \cite{zhang2018context}, Filter Response Normalization (FRN) \cite{singh2019filter}, or Cross-Iteration Batch Normalization (CBN) \cite{yao2020cross}
	\item \textbf{Skip-connections}: Residual connections, Weighted residual connections, Multi-input weighted residual connections, or Cross stage partial connections (CSP)
\end{itemize}

As for training activation function, since PReLU and SELU are more difficult to train, and ReLU6 is specifically designed for quantization network, we therefore remove the above activation functions from the candidate list. In the method of reqularization, the people who published DropBlock have compared their method with other methods in detail, and their regularization method has won a lot. Therefore, we did not hesitate to choose DropBlock as our regularization method. As for the selection of normalization method, since we focus on a training strategy that uses only one GPU, syncBN is not considered.

\subsection{Additional improvements}

In order to make the designed detector more suitable for training on single GPU, we made additional design and improvement as follows:

\begin{itemize}
	\item We introduce a new method of data augmentation – Mosaic, and Self-Adversarial Training (SAT)
	\item We select optimal hyper-parameters while applying genetic algorithms
	\item We modify some exsiting methods to make our design suitble for efficient training and detection - modified SAM, modified PAN, and Cross mini-Batch Normalization (CmBN)
\end{itemize}

\begin{figure}[h]
	\begin{center}
		\includegraphics[width=0.99\linewidth]{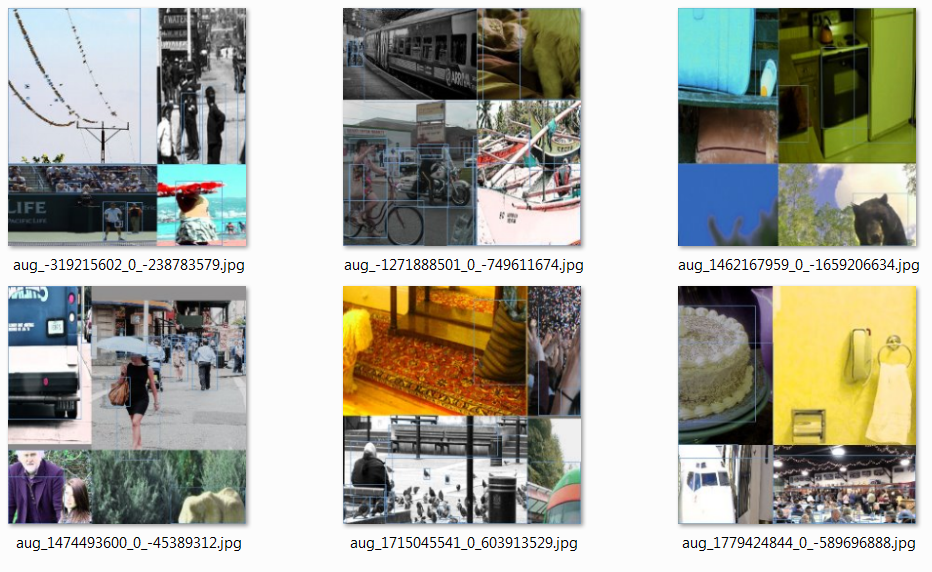}
	\end{center}
    \vspace{-6mm}
	\caption{Mosaic represents a new method of data augmentation.  }
	\label{fig:mosaic}
\end{figure}

Mosaic represents a new data augmentation method that mixes 4 training images. Thus 4 different contexts are mixed, while CutMix mixes only 2 input images. This allows detection of objects outside their normal context. In addition, batch normalization calculates activation statistics from 4 different images on each layer. This significantly reduces the need for a large mini-batch size.

Self-Adversarial Training (SAT) also represents a new data augmentation technique that operates in 2 forward – backward stages. In the 1st stage the neural network alters the original image instead of the network weights. In this way the neural network executes an adversarial attack on itself, altering the original image to create the deception that there is no desired object on the image. In the 2nd stage, the neural network is trained to detect an object on this modified image in the normal way.

\begin{figure}[h]
	\begin{center}
		\includegraphics[width=0.99\linewidth]{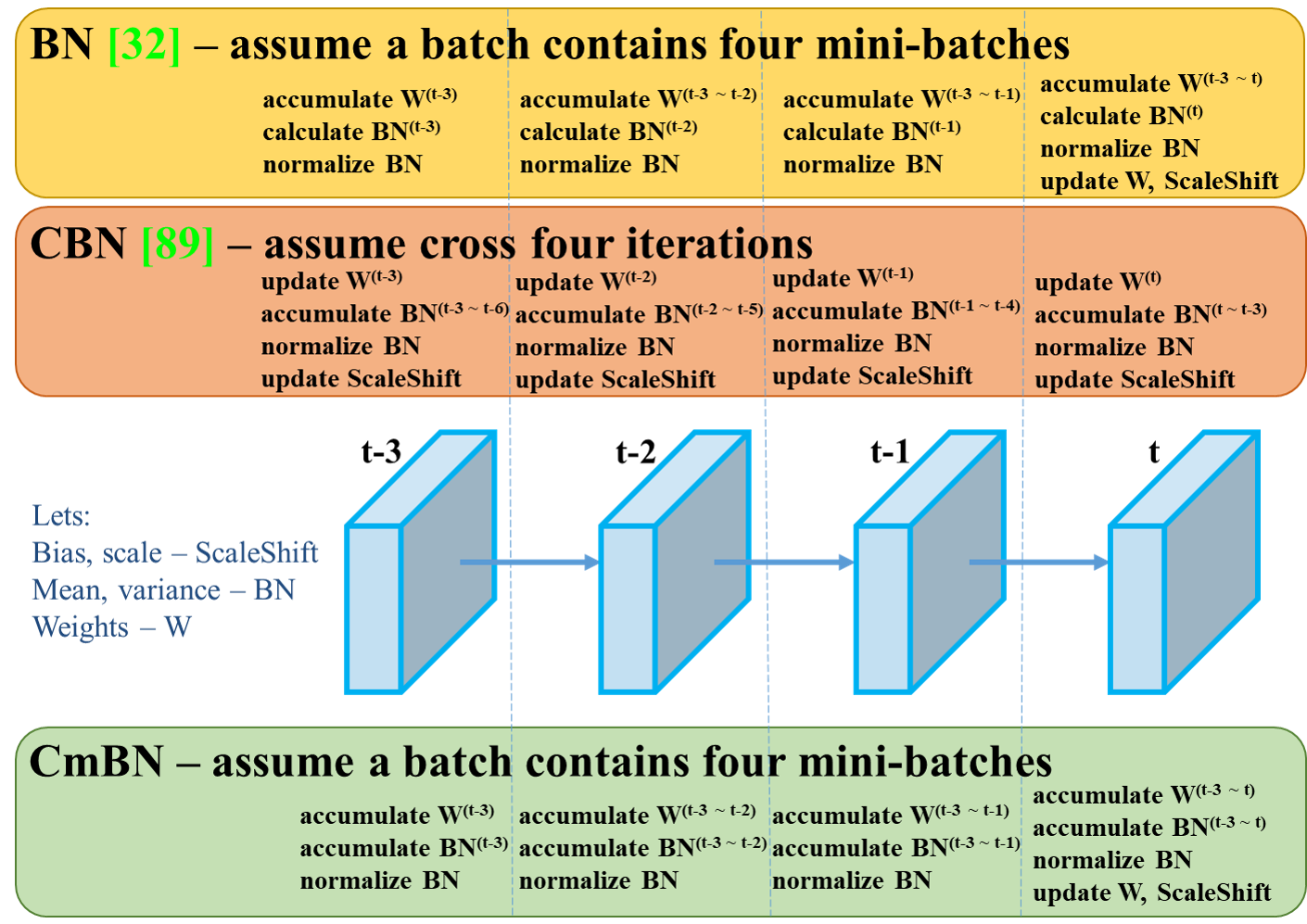}
	\end{center}
    \vspace{-6mm}
	\caption{Cross mini-Batch Normalization.  }
	\label{fig:cmbn}
\end{figure}

CmBN represents a CBN modified version, as shown in Figure \ref{fig:cmbn}, defined as Cross mini-Batch Normalization (CmBN). This collects statistics only between mini-batches within a single batch.

\begin{figure}[h]
	\begin{center}
		\includegraphics[width=0.9\linewidth]{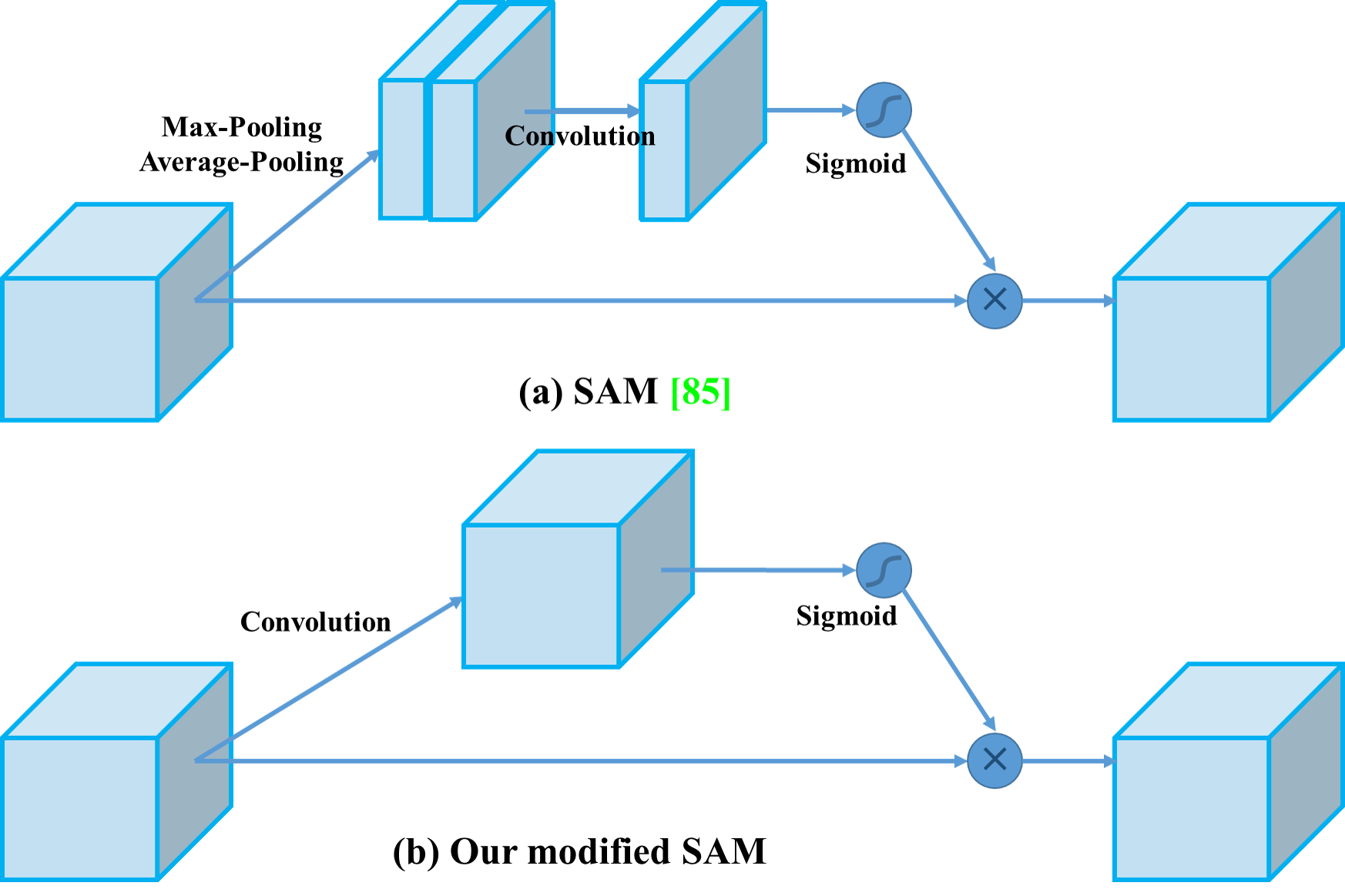}
	\end{center}
    \vspace{-6mm}
	\caption{Modified SAM.  }
	\label{fig:sam}
\end{figure}

\begin{figure}[h]
	\begin{center}
		\includegraphics[width=0.9\linewidth]{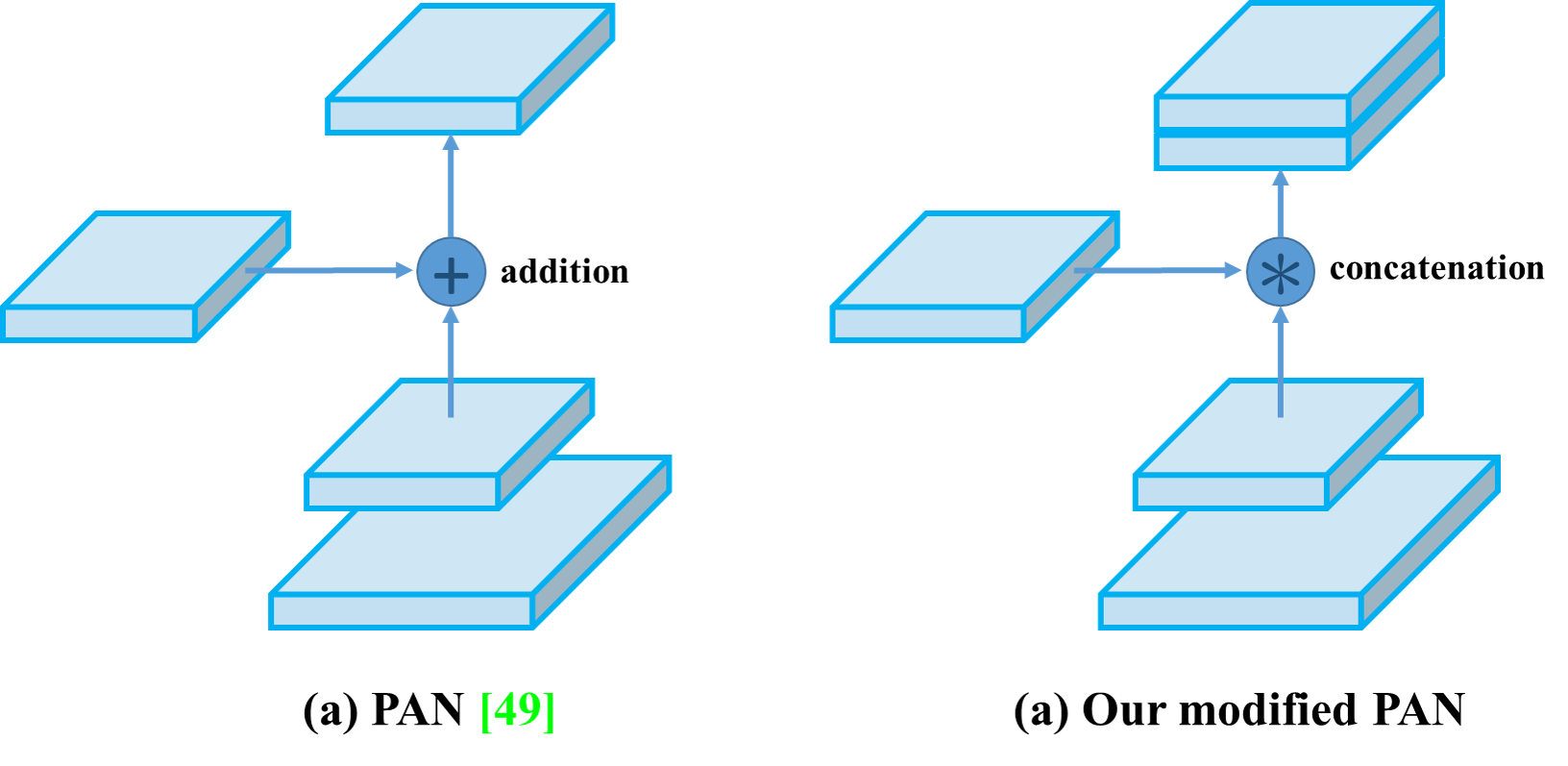}
	\end{center}
    \vspace{-6mm}
	\caption{Modified PAN.  }
	\label{fig:pan}
\end{figure}

We modify SAM from spatial-wise attention to point-wise attention, and replace shortcut connection of PAN to concatenation, as shown in Figure \ref{fig:sam} and Figure \ref{fig:pan}, respectively.

\subsection{YOLOv4}

In this section, we shall elaborate the details of YOLOv4.

\vspace{4mm}

\textbf{YOLOv4 consists of}:

\begin{itemize}
	\item Backbone: CSPDarknet53 \cite{wang2020cspnet} \\
	\vspace{-2mm}
	\item Neck: SPP \cite{he2015spatial}, PAN \cite{liu2018path} \\
	\vspace{-2mm}
	\item Head: YOLOv3 \cite{redmon2018yolov3} \\
\end{itemize}

\textbf{YOLO v4 uses}:

\begin{itemize}
	\item Bag of Freebies (BoF) for backbone: CutMix and Mosaic data augmentation, DropBlock regularization, Class label smoothing
	\item Bag of Specials (BoS) for backbone: Mish activation, Cross-stage partial connections (CSP), Multi-input weighted residual connections (MiWRC)
	\item Bag of Freebies (BoF) for detector: CIoU-loss, CmBN, DropBlock regularization, Mosaic data augmentation, Self-Adversarial Training, Eliminate grid sensitivity, Using multiple anchors for a single ground truth, Cosine annealing scheduler \cite{loshchilov2016sgdr}, Optimal hyper-parameters, Random training shapes
	\item Bag of Specials (BoS) for detector: Mish activation, SPP-block, SAM-block, PAN  path-aggregation block, DIoU-NMS
\end{itemize}


\section{Experiments}

We test the influence of different training improvement techniques on accuracy of the classifier on ImageNet (ILSVRC 2012 val) dataset, and then on the accuracy of the detector on MS COCO (test-dev 2017) dataset.

\subsection{Experimental setup}

In ImageNet image classification experiments, the default hyper-parameters are as follows: the training steps is 8,000,000; the batch size and the mini-batch size are 128 and 32, respectively; the polynomial decay learning rate scheduling strategy is adopted with initial learning rate 0.1; the warm-up steps is 1000; the momentum and weight decay are respectively set as 0.9 and 0.005. All of our BoS experiments use the same hyper-parameter as the default setting, and in the BoF experiments, we add an additional 50\% training steps. In the BoF experiments, we verify MixUp, CutMix, Mosaic, Bluring data augmentation, and label smoothing regularization methods. In the BoS experiments, we compared the effects of LReLU, Swish, and Mish activation function. All experiments are trained with a 1080 Ti or 2080 Ti GPU.

In MS COCO object detection experiments, the default hyper-parameters are as follows: the training steps is 500,500; the step decay learning rate scheduling strategy is adopted with initial learning rate 0.01 and multiply with a factor 0.1 at the 400,000 steps and the 450,000 steps, respectively; The momentum and weight decay are respectively set as 0.9 and 0.0005.  All architectures use a single GPU to execute multi-scale training in the batch size of 64 while mini-batch size is 8 or 4 depend on the architectures and GPU memory limitation. Except for using genetic algorithm for hyper-parameter search experiments, all other experiments use default setting. Genetic algorithm used YOLOv3-SPP to train with GIoU loss and search 300 epochs for min-val 5k sets. We adopt searched learning rate 0.00261, momentum 0.949, IoU threshold for assigning ground truth 0.213, and loss normalizer 0.07 for genetic algorithm experiments. We have verified a large number of BoF, including grid sensitivity elimination, mosaic data augmentation, IoU threshold, genetic algorithm, class label smoothing, cross mini-batch normalization, self-adversarial training, cosine annealing scheduler, dynamic mini-batch size, DropBlock, Optimized Anchors, different kind of IoU losses. We also conduct experiments on various BoS, including Mish, SPP, SAM, RFB, BiFPN, and Gaussian YOLO \cite{choi2019gaussian}. For all experiments, we only use one GPU for training, so techniques such as syncBN that optimizes multiple GPUs are not used.

\subsection{Influence of different features on Classifier training}

\begin{figure}[h]
	\begin{center}
		\includegraphics[width=0.99\linewidth]{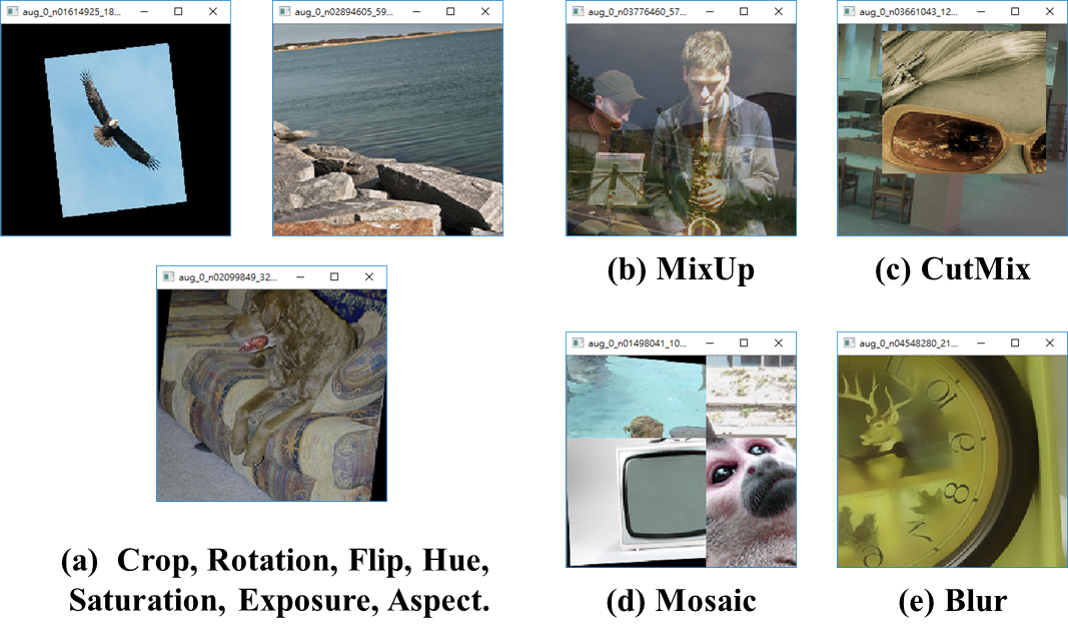}
	\end{center}
	\caption{Various method of data augmentation.  }
	\label{fig:aug}
\end{figure}

First, we study the influence of different features on classifier training; specifically, the influence of Class label smoothing, the influence of different data augmentation techniques, bilateral blurring, MixUp, CutMix and Mosaic, as shown in Fugure \ref{fig:aug}, and the influence of different activations, such as Leaky-ReLU (by default), Swish, and Mish.

In our experiments, as illustrated in Table \ref{table:cx50}, the classifier's accuracy is improved by introducing the features such as: CutMix and Mosaic data augmentation, Class label smoothing, and Mish activation. As a result, our BoF-backbone (Bag of Freebies) for classifier training includes the following: CutMix and Mosaic data augmentation and Class label smoothing. In addition we use Mish activation as a complementary option, as shown in Table \ref{table:cx50} and Table \ref{table:cd53}.

\begin{table}[h]
	\centering
	\begin{threeparttable}[h]
		\footnotesize
		\caption{Influence of BoF and Mish on the CSPResNeXt-50 classifier accuracy.}
		\label{table:cx50}
		\setlength\tabcolsep{1.0pt}
		\begin{tabular}{lccccccccc}
			\toprule
			 & \textbf{MixUp} & \textbf{CutMix} & \textbf{Mosaic} & \textbf{Bluring} & \begin{tabular}{@{}c@{}}\textbf{Label} \\ \textbf{Smoothing}\end{tabular} & \textbf{Swish} & \textbf{Mish} & \textbf{Top-1} & \textbf{Top-5} \\	
			\midrule
			 &  &  &  &  &  &  &  & 77.9\% & 94.0\% \\
			 & $\checkmark$ &  &  &  &  &  &  & 77.2\% & \textbf{94.0\%} \\
			 &  & $\checkmark$ &  &  &  &  &  & \textbf{78.0\%} & \textbf{94.3\%} \\
			 &  &  & $\checkmark$ &  &  &  &  & \textbf{78.1\%} & \textbf{94.5\%} \\
			 &  &  &  & $\checkmark$ &  &  &  & 77.5\% & 93.8\% \\
			 &  &  &  &  & $\checkmark$ &  &  & \textbf{78.1\%} & \textbf{94.4\%} \\
			 &  &  &  &  &  & $\checkmark$ &  & 64.5\% & 86.0\% \\
			 &  &  &  &  &  &  & $\checkmark$ & \textbf{78.9\%} & \textbf{94.5\%} \\
			 &  & $\checkmark$ & $\checkmark$ &  & $\checkmark$ &  &  & \textbf{78.5\%} & \textbf{94.8\%} \\
			 &  & $\checkmark$ & $\checkmark$ &  & $\checkmark$ &  & $\checkmark$ & \textbf{79.8\%} & \textbf{95.2\%} \\
			\bottomrule
		\end{tabular}
	\end{threeparttable}
\end{table}

\begin{table}[h]
\centering
\begin{threeparttable}[h]
	\footnotesize
	\caption{Influence of BoF and Mish on the CSPDarknet-53 classifier accuracy.}
	\label{table:cd53}
	\setlength\tabcolsep{1.0pt}
	\begin{tabular}{lccccccccc}
		\toprule
		& \textbf{MixUp} & \textbf{CutMix} & \textbf{Mosaic} & \textbf{Bluring} & \begin{tabular}{@{}c@{}}\textbf{Label} \\ \textbf{Smoothing}\end{tabular} & \textbf{Swish} & \textbf{Mish} & \textbf{Top-1} & \textbf{Top-5} \\		
		\midrule
		 &  &  &  &  &  &  &  & 77.2\% & 93.6\% \\
		 &  & $\checkmark$ & $\checkmark$ &  & $\checkmark$ &  &  & \textbf{77.8\%} & \textbf{94.4\%} \\
		 &  & $\checkmark$ & $\checkmark$ &  & $\checkmark$ &  & $\checkmark$ & \textbf{78.7\%} & \textbf{94.8\%} \\
		\bottomrule
	\end{tabular}
\end{threeparttable}
\end{table}

\subsection{Influence of different features on Detector training}

Further study concerns the influence of different Bag-of-Freebies (BoF-detector) on the detector training accuracy, as shown in Table \ref{table:bof}. We significantly expand the BoF list through studying different features that increase the detector accuracy without affecting FPS:

\begin{table*}[h]
	\centering
	\begin{threeparttable}[h]
		\footnotesize
		\caption{Ablation Studies of Bag-of-Freebies. (CSPResNeXt50-PANet-SPP, 512x512).}
		\label{table:bof}
		\begin{tabular}{lcccccccccccc}
			\toprule
			\textbf{S} & \textbf{M} & \textbf{IT} & \textbf{GA} & \textbf{LS} & \textbf{CBN} & \textbf{CA} & \textbf{DM} & \textbf{OA} & \textbf{loss} & \textbf{AP} & \textbf{AP$_{50}$} & \textbf{AP$_{75}$} \\	
			\midrule
			&  &  &  &  &  &  &  &  & MSE & 38.0\% & 60.0\% & 40.8\% \\
			$\checkmark$ &  &  &  &  &  &  &  &  & MSE & 37.7\% & 59.9\% & 40.5\% \\
			& $\checkmark$ &  &  &  &  &  &  &  & MSE & \textbf{39.1\%} & \textbf{61.8\%} & \textbf{42.0\%} \\
			&  & $\checkmark$ &  &  &  &  &  &  & MSE & 36.9\% & 59.7\% & 39.4\% \\
			&  &  & $\checkmark$ &  &  &  &  &  & MSE & \textbf{38.9\%} & \textbf{61.7\%} & \textbf{41.9\%} \\
			&  &  &  & $\checkmark$ &  &  &  &  & MSE & 33.0\% & 55.4\% & 35.4\% \\
			&  &  &  &  & $\checkmark$ &  &  &  & MSE & \textbf{38.4\%} & \textbf{60.7\%} & \textbf{41.3\%} \\
			&  &  &  &  &  & $\checkmark$ &  &  & MSE & \textbf{38.7\%} & \textbf{60.7\%} & \textbf{41.9\%} \\
			&  &  &  &  &  &  & $\checkmark$ &  & MSE & 35.3\% & 57.2\% & 38.0\% \\
			$\checkmark$ &  &  &  &  &  &  &  &  & GIoU & \textbf{39.4\%} & 59.4\% & \textbf{42.5\%} \\
			$\checkmark$ &  &  &  &  &  &  &  &  & DIoU & \textbf{39.1\%} & 58.8\% & \textbf{42.1\%} \\
			$\checkmark$ &  &  &  &  &  &  &  &  & CIoU & \textbf{39.6\% }& 59.2\% & \textbf{42.6\%} \\
			$\checkmark$ & $\checkmark$ & $\checkmark$ & $\checkmark$ &  &  &  &  &  & CIoU & \textbf{41.5\%} & \textbf{64.0\%} & \textbf{44.8\%} \\
			& $\checkmark$ &  & $\checkmark$ &  &  &  &  & $\checkmark$ & CIoU & 36.1\% & 56.5\% & 38.4\% \\
			$\checkmark$ & $\checkmark$ & $\checkmark$ & $\checkmark$ &  &  &  &  & $\checkmark$ & MSE & \textbf{40.3\%} & \textbf{64.0\%} & \textbf{43.1\%} \\
			$\checkmark$ & $\checkmark$ & $\checkmark$ & $\checkmark$ &  &  &  &  & $\checkmark$ & GIoU & \textbf{42.4\%} & \textbf{64.4\%} & \textbf{45.9\%} \\
			$\checkmark$ & $\checkmark$ & $\checkmark$ & $\checkmark$ &  &  &  &  & $\checkmark$ & CIoU & \textbf{42.4\%} & \textbf{64.4\%} & \textbf{45.9\%} \\
			\bottomrule
		\end{tabular}
	\end{threeparttable}
\end{table*}

\begin{itemize}
\item S: Eliminate grid sensitivity – the equation $b_x = \sigma(t_x) + c_x, b_y = \sigma(t_y) + c_y$, where $c_x$ and $c_y$ are always whole numbers, is used in YOLOv3 for evaluating the object coordinates, therefore, extremely high $t_x$ absolute values are required for the $b_x$ value approaching the $c_x$ or $c_x+1$ values.  We solve this problem through multiplying the sigmoid by a factor exceeding 1.0, so eliminating the effect of grid on which the object is undetectable.
\item M: Mosaic data augmentation - using the 4-image mosaic during training instead of single image
\item IT: IoU threshold - using multiple anchors for a single ground truth IoU (truth, anchor) $>$ IoU\_threshold
\item GA: Genetic algorithms - using genetic algorithms for selecting the optimal hyperparameters during network training on the first 10\% of time periods
\item LS: Class label smoothing - using class label smoothing for sigmoid activation
\item CBN: CmBN - using Cross mini-Batch Normalization for collecting statistics inside the entire batch, instead of collecting statistics inside a single mini-batch
\item CA: Cosine annealing scheduler - altering the learning rate during sinusoid training
\item DM: Dynamic mini-batch size - automatic increase of mini-batch size during small resolution training by using Random training shapes 
\item OA: Optimized Anchors - using the optimized anchors for training with the 512x512 network resolution
\item GIoU, CIoU, DIoU, MSE - using different loss algorithms for bounded box regression
\end{itemize}

Further study concerns the influence of different Bag-of-Specials (BoS-detector) on the detector training accuracy, including PAN, RFB, SAM, Gaussian YOLO (G), and ASFF, as shown in Table \ref{table:bos}. In our experiments, the detector gets best performance when using SPP, PAN, and SAM.

\begin{table}[h]
	\centering
	\begin{threeparttable}[h]
		\footnotesize
		\caption{Ablation Studies of Bag-of-Specials. (Size 512x512).}
		\label{table:bos}
		\setlength\tabcolsep{3.5pt}
		\begin{tabular}{lccc}
			\toprule
			\textbf{Model} & \textbf{AP} & \textbf{AP$_{50}$} & \textbf{AP$_{75}$} \\			
			\midrule
			CSPResNeXt50-PANet-SPP & 42.4\% & 64.4\% & 45.9\% \\
			CSPResNeXt50-PANet-SPP-RFB & 41.8\% & 62.7\% & 45.1\% \\
			CSPResNeXt50-PANet-SPP-SAM & \textbf{42.7\%} & \textbf{64.6\%} & \textbf{46.3\%} \\
			CSPResNeXt50-PANet-SPP-SAM-G & 41.6\% & 62.7\% & 45.0\% \\
			CSPResNeXt50-PANet-SPP-ASFF-RFB & 41.1\% & 62.6\% & 44.4\% \\
			\bottomrule
		\end{tabular}
	\end{threeparttable}
\end{table}

\subsection{Influence of different backbones and pre-trained weightings on Detector training}

Further on we study the influence of different backbone models on the detector accuracy, as shown in Table \ref{table:pretrain}. We notice that the model characterized with the best classification accuracy is not always the best in terms of the detector accuracy. 

First, although classification accuracy of CSPResNeXt-50 models trained with different features is higher compared to CSPDarknet53 models, the CSPDarknet53 model shows higher accuracy in terms of object detection. 

Second, using BoF and Mish for the CSPResNeXt50 classifier training increases its classification accuracy, but further application of these pre-trained weightings for detector training reduces the detector accuracy. However, using BoF and Mish for the CSPDarknet53 classifier training increases the accuracy of both the classifier and the detector which uses this classifier pre-trained weightings. The net result is that backbone CSPDarknet53 is more suitable for the detector than for CSPResNeXt50.

We observe that the CSPDarknet53 model demonstrates a greater ability to increase the detector accuracy owing to various improvements.

\begin{table}[h]
\centering
\begin{threeparttable}[h]
	\footnotesize
	\caption{Using different classifier pre-trained weightings for detector training (all other training parameters are similar in all models) .}
	\label{table:pretrain}
	\begin{tabular}{lcccc}
		\toprule
		\textbf{Model (with optimal setting)}  & \textbf{Size} & \textbf{AP} & \textbf{AP$_{50}$} & \textbf{AP$_{75}$} \\			
		\midrule
		\textbf{CSPResNeXt50-PANet-SPP} & 512x512 & 42.4 & 64.4 & 45.9 \\	
		\begin{tabular}{@{}l@{}}\textbf{CSPResNeXt50-PANet-SPP} \\ (BoF-backbone)\end{tabular} & 512x512 & 42.3 & 64.3 & 45.7 \\	
		\begin{tabular}{@{}l@{}}\textbf{CSPResNeXt50-PANet-SPP} \\ (BoF-backbone + Mish)\end{tabular} & 512x512 & 42.3 & 64.2 & 45.8 \\	
		\midrule
		\begin{tabular}{@{}l@{}}\textbf{CSPDarknet53-PANet-SPP} \\ (BoF-backbone)\end{tabular} & 512x512 & 42.4 & 64.5 & 46.0 \\	
		\begin{tabular}{@{}l@{}}\textbf{CSPDarknet53-PANet-SPP} \\ (BoF-backbone + Mish)\end{tabular} & 512x512 & 43.0 & 64.9 & 46.5 \\		
		\bottomrule
	\end{tabular}
\end{threeparttable}
\end{table}

\subsection{Influence of different mini-batch size on Detector training}

Finally, we analyze the results obtained with models trained with different mini-batch sizes, and the results are shown in Table \ref{table:bs}. From the results shown in Table \ref{table:bs}, we found that after adding BoF and BoS training strategies, the mini-batch size has almost no effect on the detector's performance. This result shows that after the introduction of BoF and BoS, it is no longer necessary to use expensive GPUs for training. In other words, anyone can use only a conventional GPU to train an excellent detector.

\begin{table}[h]
	\centering
	\begin{threeparttable}[h]
		\footnotesize
		\caption{Using different mini-batch size for detector training.}
		\label{table:bs}
		\begin{tabular}{lcccc}
			\toprule
			\textbf{Model (without OA)}  & \textbf{Size} & \textbf{AP} & \textbf{AP$_{50}$} & \textbf{AP$_{75}$} \\			
			\midrule	
			\begin{tabular}{@{}l@{}}\textbf{CSPResNeXt50-PANet-SPP} \\ (without BoF/BoS, mini-batch 4)\end{tabular} & 608 & 37.1 & 59.2 & 39.9 \\	
			\begin{tabular}{@{}l@{}}\textbf{CSPResNeXt50-PANet-SPP} \\  (without BoF/BoS, mini-batch 8)\end{tabular} & 608 & 38.4 & 60.6 & 41.6 \\	
			\midrule
			\begin{tabular}{@{}l@{}}\textbf{CSPDarknet53-PANet-SPP} \\  (with BoF/BoS, mini-batch 4)\end{tabular} & 512 & 41.6 & 64.1 & 45.0 \\	
			\begin{tabular}{@{}l@{}}\textbf{CSPDarknet53-PANet-SPP} \\  (with BoF/BoS, mini-batch 8)\end{tabular} & 512 & 41.7 & 64.2 & 45.2 \\		
			\bottomrule
		\end{tabular}
	\end{threeparttable}
\end{table}


\begin{figure*}[t]
	\begin{center}
		\includegraphics[width=0.8\linewidth]{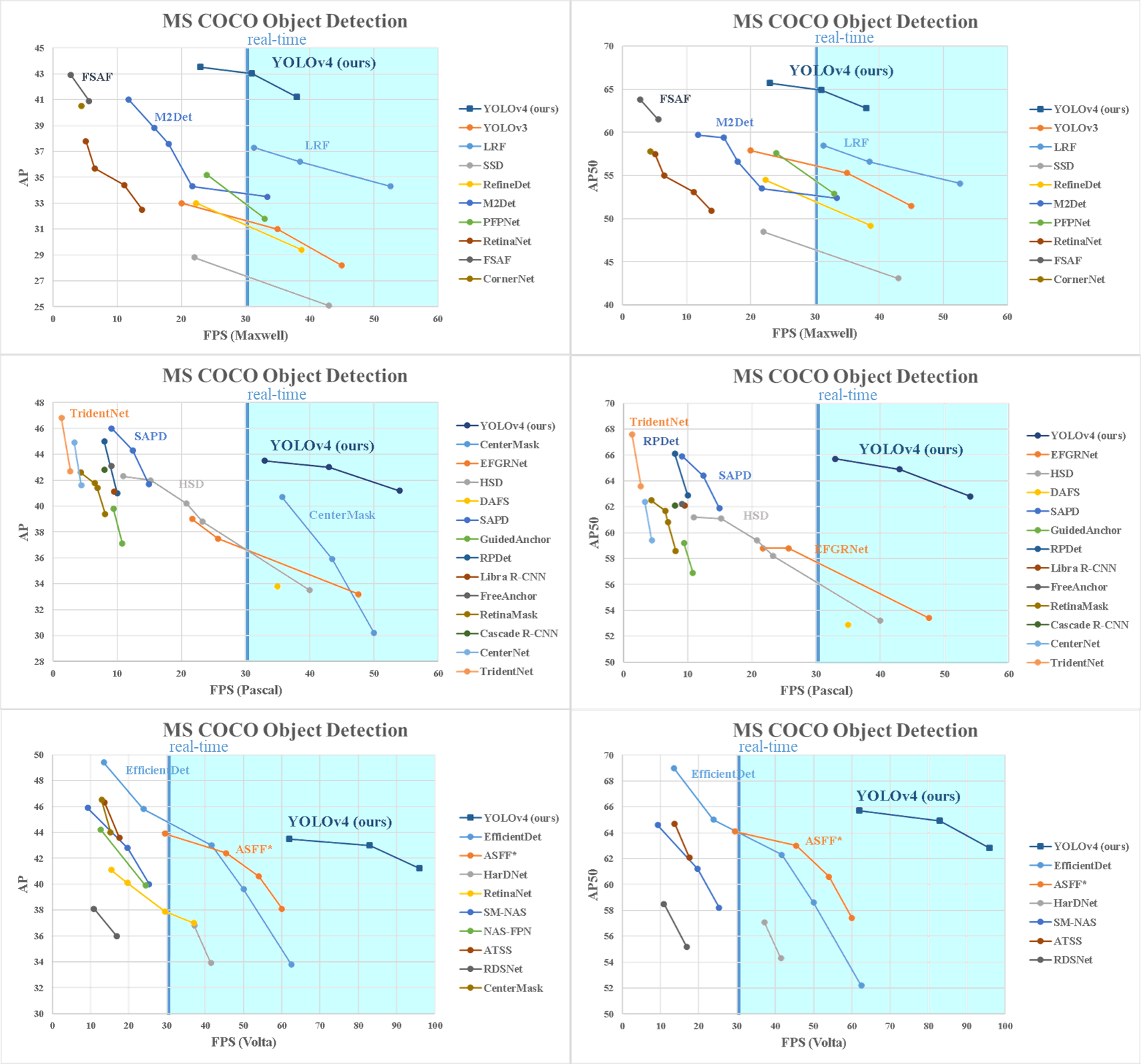}
	\end{center}
	\vspace{-4mm}
	\caption{Comparison of the speed and accuracy of different object detectors. (Some articles stated the FPS of their detectors for only one of the GPUs: Maxwell/Pascal/Volta) }
	\label{fig:fps}
\end{figure*}

\section{Results}

Comparison of the results obtained with other state-of-the-art object detectors are shown in Figure \ref{fig:fps}. Our YOLOv4 are located on the Pareto optimality curve and are superior to the fastest and most accurate detectors in terms of both speed and accuracy.

Since different methods use GPUs of different architectures for inference time verification, we operate YOLOv4 on commonly adopted GPUs of Maxwell, Pascal, and Volta architectures, and compare them with other state-of-the-art methods. Table \ref{table:maxwell} lists the frame rate comparison results of using Maxwell GPU, and it can be GTX Titan X (Maxwell) or Tesla M40 GPU. Table \ref{table:pascal} lists the frame rate comparison results of using Pascal GPU, and it can be Titan X (Pascal), Titan Xp, GTX 1080 Ti, or Tesla P100 GPU. As for Table \ref{table:volta}, it lists the frame rate comparison results of using Volta GPU, and it can be Titan Volta or Tesla V100 GPU.

\begin{table*}[h]
	\centering
	\begin{threeparttable}[h]
		\footnotesize
		\caption{Comparison of the speed and accuracy of different object detectors on the MS COCO dataset (test-dev 2017). (Real-time detectors with FPS 30 or higher are highlighted here. We compare the results with batch=1 without using tensorRT.)}
		\label{table:maxwell}
		\begin{tabular}{llcccccccc}
			\toprule
			\textbf{Method} & \textbf{Backbone} & \textbf{Size} & \textbf{FPS} & \textbf{AP} & \textbf{AP$_{50}$} & \textbf{AP$_{75}$} & \textbf{AP$_S$} & \textbf{AP$_M$} & \textbf{AP$_L$} \\	
			\midrule
			\multicolumn{10}{c}{\textbf{YOLOv4: Optimal Speed and Accuracy of Object Detection}} \\
			\rowcolor{cyan}\textbf{YOLOv4} & CSPDarknet-53 & 416 & 38 (M) & 41.2\% & 62.8\% & 44.3\% & 20.4\% & 44.4\% & 56.0\% \\
			\rowcolor{cyan}\textbf{YOLOv4} & CSPDarknet-53 & 512 & 31 (M) & \textbf{43.0\%} & \textbf{64.9\%} & \textbf{46.5\%} & \textbf{24.3\%} & \textbf{46.1\%} & \textbf{55.2\%} \\
			\textbf{YOLOv4} & CSPDarknet-53 & 608 & 23 (M) & 43.5\% & 65.7\% & 47.3\% & 26.7\% & 46.7\% & 53.3\% \\
			\midrule
			\multicolumn{10}{c}{\textbf{Learning Rich Features at High-Speed for Single-Shot Object Detection \cite{wang2019learning}}} \\
			\rowcolor{cyan}LRF & VGG-16 & 300 & 76.9 (M) & 32.0\% & 51.5\% & 33.8\% & 12.6\% & 34.9\% & 47.0\% \\
			\rowcolor{cyan}LRF & ResNet-101 & 300 & 52.6 (M) & 34.3\% & 54.1\% & 36.6\% & 13.2\% & 38.2\% & 50.7\% \\
			\rowcolor{cyan}LRF & VGG-16 & 512 & 38.5 (M) & 36.2\% & 56.6\% & 38.7\% & 19.0\% & 39.9\% & 48.8\% \\
			\rowcolor{cyan}LRF & ResNet-101 & 512 & 31.3 (M) & 37.3\% & 58.5\% & 39.7\% & 19.7\% & 42.8\% & 50.1\% \\
			\midrule
			\multicolumn{10}{c}{\textbf{Receptive Field Block Net for Accurate and Fast Object Detection \cite{liu2018receptive}}} \\
			\rowcolor{cyan}RFBNet & VGG-16 & 300 & 66.7 (M) & 30.3\% & 49.3\% & 31.8\% & 11.8\% & 31.9\% & 45.9\% \\
			\rowcolor{cyan}RFBNet & VGG-16 & 512 & 33.3 (M) & 33.8\% & 54.2\% & 35.9\% & 16.2\% & 37.1\% & 47.4\% \\
			\rowcolor{cyan}RFBNet-E & VGG-16 & 512 & 30.3 (M) & 34.4\% & 55.7\% & 36.4\% & 17.6\% & 37.0\% & 47.6\% \\
			\midrule
			\multicolumn{10}{c}{\textbf{YOLOv3: An incremental improvement} \cite{redmon2018yolov3}} \\
			\rowcolor{cyan}YOLOv3 & Darknet-53 & 320 & 45 (M) & 28.2\% & 51.5\% & 29.7\% & 11.9\% & 30.6\% & 43.4\% \\
			\rowcolor{cyan}YOLOv3 & Darknet-53 & 416 & 35 (M) & 31.0\% & 55.3\% & 32.3\% & 15.2\% & 33.2\% & 42.8\% \\
			YOLOv3 & Darknet-53 & 608 & 20 (M) & 33.0\% & 57.9\% & 34.4\% & 18.3\% & 35.4\% & 41.9\% \\
			YOLOv3-SPP & Darknet-53 & 608 & 20 (M) & 36.2\% & 60.6\% & 38.2\% & 20.6\% & 37.4\% & 46.1\% \\
			\midrule
			\multicolumn{10}{c}{\textbf{SSD: Single shot multibox detector} \cite{liu2016ssd}} \\
			\rowcolor{cyan}SSD & VGG-16 & 300 & 43 (M) & 25.1\% & 43.1\% & 25.8\% & 6.6\% & 25.9\% & 41.4\% \\
			SSD & VGG-16 & 512 & 22 (M) & 28.8\% & 48.5\% & 30.3\% & 10.9\% & 31.8\% & 43.5\% \\
			\midrule
			\multicolumn{10}{c}{\textbf{Single-shot refinement neural network for object detection} \cite{zhang2018single}} \\
			\rowcolor{cyan}RefineDet & VGG-16 & 320 & 38.7 (M) & 29.4\% & 49.2\% & 31.3\% & 10.0\% & 32.0\% & 44.4\% \\
			RefineDet & VGG-16 & 512 & 22.3 (M) & 33.0\% & 54.5\% & 35.5\% & 16.3\% & 36.3\% & 44.3\% \\
			\midrule
			\multicolumn{10}{c}{\textbf{M2det: A single-shot object detector based on multi-level feature pyramid network} \cite{zhao2019m2det}} \\
			\rowcolor{cyan}M2det & VGG-16 & 320 & 33.4 (M) & 33.5\% & 52.4\% & 35.6\% & 14.4\% & 37.6\% & 47.6\% \\
			M2det & ResNet-101 & 320 & 21.7 (M) & 34.3\% & 53.5\% & 36.5\% & 14.8\% & 38.8\% & 47.9\% \\
			M2det & VGG-16 & 512 & 18 (M) & 37.6\% & 56.6\% & 40.5\% & 18.4\% & 43.4\% & 51.2\% \\
			M2det & ResNet-101 & 512 & 15.8 (M) & 38.8\% & 59.4\% & 41.7\% & 20.5\% & 43.9\% & 53.4\% \\
			M2det & VGG-16 & 800 & 11.8 (M) & 41.0\% & 59.7\% & 45.0\% & 22.1\% & 46.5\% & 53.8\% \\
			\midrule
			\multicolumn{10}{c}{\textbf{Parallel Feature Pyramid Network for Object Detection} \cite{kim2018parallel}} \\
			\rowcolor{cyan}PFPNet-R & VGG-16 & 320 & 33 (M) & 31.8\% & 52.9\% & 33.6\% & 12\% & 35.5\% & 46.1\% \\
			PFPNet-R & VGG-16 & 512 & 24 (M) & 35.2\% & 57.6\% & 37.9\% & 18.7\% & 38.6\% & 45.9\% \\
			\midrule
			\midrule
			\multicolumn{10}{c}{\textbf{Focal Loss for Dense Object Detection} \cite{lin2017focal}} \\
			RetinaNet & ResNet-50 & 500 & 13.9 (M) & 32.5\% & 50.9\% & 34.8\% & 13.9\% & 35.8\% & 46.7\% \\
			RetinaNet & ResNet-101 & 500 & 11.1 (M) & 34.4\% & 53.1\% & 36.8\% & 14.7\% & 38.5\% & 49.1\% \\
			RetinaNet & ResNet-50 & 800 & 6.5 (M) & 35.7\% & 55.0\% & 38.5\% & 18.9\% & 38.9\% & 46.3\% \\
			RetinaNet & ResNet-101 & 800 & 5.1 (M) & 37.8\% & 57.5\% & 40.8\% & 20.2\% & 41.1\% & 49.2\% \\
			\midrule
			\multicolumn{10}{c}{\textbf{Feature Selective Anchor-Free Module for Single-Shot Object Detection \cite{zhu2019feature}}} \\
			AB+FSAF & ResNet-101 & 800 & 5.6 (M) & 40.9\% & 61.5\% & 44.0\% & 24.0\% & 44.2\% & 51.3\% \\
			AB+FSAF & ResNeXt-101 & 800 & 2.8 (M) & 42.9\% & 63.8\% & 46.3\% & 26.6\% & 46.2\% & 52.7\% \\
			\midrule
			\multicolumn{10}{c}{\textbf{CornerNet: Detecting objects as paired keypoints} \cite{law2018cornernet}} \\
			CornerNet & Hourglass & 512 & 4.4 (M) & 40.5\% & 57.8\% & 45.3\% & 20.8\% & 44.8\% & 56.7\% \\
			\bottomrule
		\end{tabular}
	\end{threeparttable}
\end{table*}

\begin{table*}[h]
	\centering
	\begin{threeparttable}[h]
		\footnotesize
		\caption{Comparison of the speed and accuracy of different object detectors on the MS COCO dataset (test-dev 2017). (Real-time detectors with FPS 30 or higher are highlighted here. We compare the results with batch=1 without using tensorRT.)}
		\label{table:pascal}
		\begin{tabular}{llcccccccc}
			\toprule
			\textbf{Method} & \textbf{Backbone} & \textbf{Size} & \textbf{FPS} & \textbf{AP} & \textbf{AP$_{50}$} & \textbf{AP$_{75}$} & \textbf{AP$_S$} & \textbf{AP$_M$} & \textbf{AP$_L$} \\	
			\midrule
			\multicolumn{10}{c}{\textbf{YOLOv4: Optimal Speed and Accuracy of Object Detection}} \\
			\rowcolor{cyan}\textbf{YOLOv4} & CSPDarknet-53 & 416 & 54 (P) & 41.2\% & 62.8\% & 44.3\% & 20.4\% & 44.4\% & 56.0\% \\
			\rowcolor{cyan}\textbf{YOLOv4} & CSPDarknet-53 & 512 & 43 (P) & 43.0\% & 64.9\% & 46.5\% & 24.3\% & 46.1\% & 55.2\% \\
			\rowcolor{cyan}\textbf{YOLOv4} & CSPDarknet-53 & 608 & 33 (P) & \textbf{43.5\%} & \textbf{65.7\%} & \textbf{47.3\%} & \textbf{26.7\%} & \textbf{46.7\%} & 53.3\% \\
			\midrule
			\multicolumn{10}{c}{\textbf{CenterMask: Real-Time Anchor-Free Instance Segmentation \cite{lee2019centermask}}} \\
			\rowcolor{cyan}CenterMask-Lite & MobileNetV2-FPN & 600$\times$ & 50.0 (P) & 30.2\% & - & - & 14.2\% & 31.9\% & 40.9\% \\
			\rowcolor{cyan}CenterMask-Lite & VoVNet-19-FPN & 600$\times$ & 43.5 (P) & 35.9\% & - & - & 19.6\% & 38.0\% & 45.9\% \\
			\rowcolor{cyan}CenterMask-Lite & VoVNet-39-FPN & 600$\times$ & 35.7 (P) & 40.7\% & - & - & 22.4\% & 43.2\% & \textbf{53.5\%} \\
			\midrule
			\multicolumn{10}{c}{\textbf{Enriched Feature Guided Refinement Network for Object Detection \cite{nie2019enriched}}} \\
			\rowcolor{cyan}EFGRNet & VGG-16 & 320 & 47.6 (P) & 33.2\% & 53.4\% & 35.4\% & 13.4\% & 37.1\% & 47.9\% \\
			EFGRNet & VG-G16 & 512 & 25.7 (P) & 37.5\% & 58.8\% & 40.4\% & 19.7\% & 41.6\% & 49.4\% \\
			EFGRNet & ResNet-101 & 512 & 21.7 (P) & 39.0\% & 58.8\% & 42.3\% & 17.8\% & 43.6\% & 54.5\% \\
			\midrule
			\multicolumn{10}{c}{\textbf{Hierarchical Shot Detector \cite{cao2019hierarchical}}} \\
			\rowcolor{cyan}HSD & VGG-16 & 320 & 40 (P) & 33.5\% & 53.2\% & 36.1\% & 15.0\% & 35.0\% & 47.8\% \\
			HSD & VGG-16 & 512 & 23.3 (P) & 38.8\% & 58.2\% & 42.5\% & 21.8\% & 41.9\% & 50.2\% \\
			HSD & ResNet-101 & 512 & 20.8 (P) & 40.2\% & 59.4\% & 44.0\% & 20.0\% & 44.4\% & 54.9\% \\
			HSD & ResNeXt-101 & 512 & 15.2 (P) & 41.9\% & 61.1\% & 46.2\% & 21.8\% & 46.6\% & 57.0\% \\
			HSD & ResNet-101 & 768 & 10.9 (P) & 42.3\% & 61.2\% & 46.9\% & 22.8\% & 47.3\% & 55.9\% \\
			\midrule
			\multicolumn{10}{c}{\textbf{Dynamic anchor feature selection for single-shot object detection} \cite{li2019dynamic}} \\
			\rowcolor{cyan} DAFS & VGG16 & 512 & 35 (P) & 33.8\% & 52.9\% & 36.9\% & 14.6\% & 37.0\% & 47.7\% \\
			\midrule
			\midrule
			\multicolumn{10}{c}{\textbf{Soft Anchor-Point Object Detection} \cite{zhu2019soft}} \\
			SAPD & ResNet-50 & - & 14.9 (P) & 41.7\% & 61.9\% & 44.6\% & 24.1\% & 44.6\% & 51.6\% \\
			SAPD & ResNet-50-DCN & - & 12.4 (P) & 44.3\% & 64.4\% & 47.7\% & 25.5\% & 47.3\% & 57.0\% \\
			SAPD & ResNet-101-DCN & - & 9.1 (P) & 46.0\% & 65.9\% & 49.6\% & 26.3\% & 49.2\% & 59.6\% \\
			\midrule
			\multicolumn{10}{c}{\textbf{Region proposal by guided anchoring} \cite{wang2019region}} \\
			RetinaNet & ResNet-50 & - & 10.8 (P) & 37.1\% & 56.9\% & 40.0\% & 20.1\% & 40.1\% & 48.0\% \\
			Faster R-CNN & ResNet-50 & - & 9.4 (P) & 39.8\% & 59.2\% & 43.5\% & 21.8\% & 42.6\% & 50.7\% \\
			\midrule
			\multicolumn{10}{c}{\textbf{RepPoints: Point set representation for object detection} \cite{yang2019reppoints}} \\
			RPDet & ResNet-101 & - & 10 (P) & 41.0\% & 62.9\% & 44.3\% & 23.6\% & 44.1\% & 51.7\% \\
			RPDet & ResNet-101-DCN & - & 8 (P) & 45.0\% & 66.1\% & 49.0\% & 26.6\% & 48.6\% & 57.5\% \\
			\midrule
			\multicolumn{10}{c}{\textbf{Libra R-CNN: Towards balanced learning for object detection} \cite{pang2019libra}} \\
			Libra R-CNN & ResNet-101 & - & 9.5 (P) & 41.1\% & 62.1\% & 44.7\% & 23.4\% & 43.7\% & 52.5\% \\
			\midrule
			\multicolumn{10}{c}{\textbf{FreeAnchor: Learning to match anchors for visual object detection} \cite{zhang2019freeanchor}} \\
			FreeAnchor & ResNet-101 & - & 9.1 (P) & 43.1\% & 62.2\% & 46.4\% & 24.5\% & 46.1\% & 54.8\% \\
			\midrule
			\multicolumn{10}{c}{\textbf{RetinaMask: Learning to Predict Masks Improves State-of-The-Art Single-Shot Detection for Free \cite{fu2019retinamask}}} \\
			RetinaMask & ResNet-50-FPN & 800$\times$ & 8.1 (P) & 39.4\% & 58.6\% & 42.3\% & 21.9\% & 42.0\% & 51.0\% \\
			RetinaMask & ResNet-101-FPN & 800$\times$ & 6.9 (P) & 41.4\% & 60.8\% & 44.6\% & 23.0\% & 44.5\% & 53.5\% \\
			RetinaMask & ResNet-101-FPN-GN & 800$\times$ & 6.5 (P) & 41.7\% & 61.7\% & 45.0\% & 23.5\% & 44.7\% & 52.8\% \\
			RetinaMask & ResNeXt-101-FPN-GN & 800$\times$ & 4.3 (P) & 42.6\% & 62.5\% & 46.0\% & 24.8\% & 45.6\% & 53.8\% \\
			\midrule
			\multicolumn{10}{c}{\textbf{Cascade R-CNN: Delving into high quality object detection} \cite{cai2018cascade}} \\
			Cascade R-CNN & ResNet-101 & - & 8 (P) & 42.8\% & 62.1\% & 46.3\% & 23.7\% & 45.5\% & 55.2\% \\
			\midrule
			\multicolumn{10}{c}{\textbf{Centernet: Object detection with keypoint triplets} \cite{duan2019centernet}} \\
			Centernet & Hourglass-52 & - & 4.4 (P) & 41.6\% & 59.4\% & 44.2\% & 22.5\% & 43.1\% & 54.1\% \\
			Centernet & Hourglass-104 & - & 3.3 (P) & 44.9\% & 62.4\% & 48.1\% & 25.6\% & 47.4\% & 57.4\% \\
			\midrule
			\multicolumn{10}{c}{\textbf{Scale-Aware Trident Networks for Object Detection} \cite{li2019scale}} \\
			TridentNet & ResNet-101 & - & 2.7 (P) & 42.7\% & 63.6\% & 46.5\% & 23.9\% & 46.6\% & 56.6\% \\
			TridentNet & ResNet-101-DCN & - & 1.3 (P) & 46.8\% & 67.6\% & 51.5\% & 28.0\% & 51.2\% & 60.5\% \\
			\bottomrule
		\end{tabular}
	\end{threeparttable}
\end{table*}

\begin{table*}[h]
	\centering
	\begin{threeparttable}[h]
		\footnotesize
		\caption{Comparison of the speed and accuracy of different object detectors on the MS COCO dataset (test-dev 2017). (Real-time detectors with FPS 30 or higher are highlighted here. We compare the results with batch=1 without using tensorRT.)}
		\label{table:volta}
		\begin{tabular}{llcccccccc}
			\toprule
			\textbf{Method} & \textbf{Backbone} & \textbf{Size} & \textbf{FPS} & \textbf{AP} & \textbf{AP$_{50}$} & \textbf{AP$_{75}$} & \textbf{AP$_S$} & \textbf{AP$_M$} & \textbf{AP$_L$} \\	
			\midrule
			\multicolumn{10}{c}{\textbf{YOLOv4: Optimal Speed and Accuracy of Object Detection}} \\
			\rowcolor{cyan}\textbf{YOLOv4} & CSPDarknet-53 & 416 & 96 (V) & 41.2\% & 62.8\% & 44.3\% & 20.4\% & 44.4\% & 56.0\% \\
			\rowcolor{cyan}\textbf{YOLOv4} & CSPDarknet-53 & 512 & 83 (V) & 43.0\% & 64.9\% & 46.5\% & 24.3\% & 46.1\% & 55.2\% \\
			\rowcolor{cyan}\textbf{YOLOv4} & CSPDarknet-53 & 608 & 62 (V) & \textbf{43.5\%} & \textbf{65.7\%} & 47.3\% & \textbf{26.7\%} & 46.7\% & 53.3\% \\
			\midrule
			\multicolumn{10}{c}{\textbf{EfficientDet: Scalable and Efficient Object Detection \cite{tan2019efficientdet}}} \\
			\rowcolor{cyan}EfficientDet-D0 & Efficient-B0 & 512 & 62.5 (V) & 33.8\% & 52.2\% & 35.8\% & 12.0\% & 38.3\% & 51.2\% \\
			\rowcolor{cyan}EfficientDet-D1 & Efficient-B1 & 640 & 50.0 (V) & 39.6\% & 58.6\% & 42.3\% & 17.9\% & 44.3\% & 56.0\% \\
			\rowcolor{cyan}EfficientDet-D2 & Efficient-B2 & 768 & 41.7 (V) & 43.0\% & 62.3\% & 46.2\% & 22.5\% & \textbf{47.0\%} & \textbf{58.4\%} \\
			EfficientDet-D3 & Efficient-B3 & 896 & 23.8 (V) & 45.8\% & 65.0\% & 49.3\% & 26.6\% & 49.4\% & 59.8\% \\
			\midrule
			\multicolumn{10}{c}{\textbf{Learning Spatial Fusion for Single-Shot Object Detection \cite{liu2019learning}}} \\
			\rowcolor{cyan}YOLOv3 + ASFF* & Darknet-53 & 320 & 60 (V) & 38.1\% & 57.4\% & 42.1\% & 16.1\% & 41.6\% & 53.6\% \\
			\rowcolor{cyan}YOLOv3 + ASFF* & Darknet-53 & 416 & 54 (V) & 40.6\% & 60.6\% & 45.1\% & 20.3\% & 44.2\% & 54.1\% \\
			\rowcolor{cyan}YOLOv3 + ASFF* & Darknet-53 & 608$\times$ & 45.5 (V) & 42.4\% & 63.0\% & \textbf{47.4\%} & 25.5\% & 45.7\% & 52.3\% \\
			YOLOv3 + ASFF* & Darknet-53 & 800$\times$ & 29.4 (V) & 43.9\% & 64.1\% & 49.2\% & 27.0\% & 46.6\% & 53.4\% \\
			\midrule
			\multicolumn{10}{c}{\textbf{HarDNet: A Low Memory Traffic Network} \cite{chao2019hardnet}} \\
			\rowcolor{cyan}RFBNet & HarDNet68 & 512 & 41.5 (V) & 33.9\% & 54.3\% & 36.2\% & 14.7\% & 36.6\% & 50.5\% \\
			\rowcolor{cyan}RFBNet & HarDNet85 & 512 & 37.1 (V) & 36.8\% & 57.1\% & 39.5\% & 16.9\% & 40.5\% & 52.9\% \\
			\midrule
			\multicolumn{10}{c}{\textbf{Focal Loss for Dense Object Detection} \cite{lin2017focal}} \\
			\rowcolor{cyan}RetinaNet & ResNet-50 & 640 & 37 (V) & 37.0\% & - & - & - & - & - \\
			RetinaNet & ResNet-101 & 640 & 29.4 (V) & 37.9\% & - & - & - & - & - \\
			RetinaNet & ResNet-50 & 1024 & 19.6 (V) & 40.1\% & - & - & - & - & - \\
			RetinaNet & ResNet-101 & 1024 & 15.4 (V) & 41.1\% & - & - & - & - & - \\
			\midrule
			\midrule
			\multicolumn{10}{c}{\textbf{SM-NAS: Structural-to-Modular Neural Architecture Search for Object Detection \cite{yao2019sm}}} \\
			SM-NAS: E2 & - & 800$\times$600 & 25.3 (V) & 40.0\% & 58.2\% & 43.4\% & 21.1\% & 42.4\% & 51.7\% \\
			SM-NAS: E3 & - & 800$\times$600 & 19.7 (V) & 42.8\% & 61.2\% & 46.5\% & 23.5\% & 45.5\% & 55.6\% \\
			SM-NAS: E5 & - & 1333$\times$800 & 9.3 (V) & 45.9\% & 64.6\% & 49.6\% & 27.1\% & 49.0\% & 58.0\% \\
			\midrule
			\multicolumn{10}{c}{\textbf{NAS-FPN: Learning scalable feature pyramid architecture for object detection} \cite{ghiasi2019fpn}} \\
			NAS-FPN & ResNet-50 & 640 & 24.4 (V) & 39.9\% & - & - & - & - & - \\
			NAS-FPN & ResNet-50 & 1024 & 12.7 (V) & 44.2\% & - & - & - & - & - \\
			\midrule
			\multicolumn{10}{c}{\textbf{Bridging the Gap Between Anchor-based and Anchor-free Detection via Adaptive Training Sample Selection \cite{zhang2019bridging}}} \\
			ATSS & ResNet-101 & 800$\times$ & 17.5 (V) & 43.6\% & 62.1\% & 47.4\% & 26.1\% & 47.0\% & 53.6\% \\
			ATSS & ResNet-101-DCN & 800$\times$ & 13.7 (V) & 46.3\% & 64.7\% & 50.4\% & 27.7\% & 49.8\% & 58.4\% \\
			\midrule
			\multicolumn{10}{c}{\textbf{RDSNet: A New Deep Architecture for Reciprocal Object Detection and Instance Segmentation \cite{wang2019rdsnet}}} \\
			RDSNet & ResNet-101 & 600 & 16.8 (V) & 36.0\% & 55.2\% & 38.7\% & 17.4\% & 39.6\% & 49.7\% \\
			RDSNet & ResNet-101 & 800 & 10.9 (V) & 38.1\% & 58.5\% & 40.8\% & 21.2\% & 41.5\% & 48.2\% \\
			\midrule
			\multicolumn{10}{c}{\textbf{CenterMask: Real-Time Anchor-Free Instance Segmentation \cite{lee2019centermask}}} \\
			CenterMask & ResNet-101-FPN & 800$\times$ & 15.2 (V) & 44.0\% & - & - & 25.8\% & 46.8\% & 54.9\% \\
			CenterMask & VoVNet-99-FPN & 800$\times$ & 12.9 (V) & 46.5\% & - & - & 28.7\% & 48.9\% & 57.2\% \\
			\bottomrule
		\end{tabular}
	\end{threeparttable}
\end{table*}


\section{Conclusions}

\vspace{-2mm}

We offer a state-of-the-art detector which is faster (FPS) and more accurate (MS COCO AP$_{50...95}$ and AP$_{50}$) than all available alternative detectors. The detector described can be trained and used on a conventional GPU with 8-16 GB-VRAM – this makes its broad use possible. The original concept of one-stage anchor-based detectors has proven its viability. We have verified a large number of features, and selected for use such of them for improving the accuracy of both the classifier and the detector. These features can be used as best-practice for future studies and developments.


\section{Acknowledgements}

\vspace{-2mm}

The authors wish to thank Glenn Jocher for the ideas of Mosaic data augmentation, the selection of hyper-parameters by using genetic algorithms and solving the grid sensitivity problem \url{https://github.com/ultralytics/yolov3}.


\clearpage
\clearpage
\clearpage
{\small
\bibliographystyle{ieee_fullname}

}
\end{document}